\documentclass{article}

\usepackage{arxiv}

\usepackage[utf8]{inputenc} 
\usepackage[T1]{fontenc}    
\usepackage{hyperref}       
\usepackage{url}            
\usepackage{booktabs}       
\usepackage{amsfonts}       
\usepackage{nicefrac}       
\usepackage{microtype}      
\usepackage{lipsum}		
\usepackage{graphicx}
\usepackage[square]{natbib}
\usepackage{doi}
\usepackage{tabularx}
\usepackage{caption}
\usepackage{subcaption}

\title{ClimateNLP: Analyzing Public Sentiment Towards Climate Change using Natural Language Processing}

\author{Ajay Krishnan T. K.\\
        Applied NLP Research Lab\\
	School of Digital Sciences\\
	Kerala University of Digital Sciences-\\
	Innovation and Technology \\
        Thiruvananthapuram, India\\
	\texttt{ajay.ds21@duk.ac.in} \\
	\And
	V. S. Anoop\\
	Applied NLP Research Lab\\
	School of Digital Sciences\\
	Kerala University of Digital Sciences-\\
	Innovation and Technology \\
        Thiruvananthapuram, India\\
	\texttt{anoop.vs@duk.ac.in} \\
}

\date{}

\renewcommand{\shorttitle}{ClimateNLP: Analyzing Public Sentiment Towards Climate Change using NLP}

\begin{document}
\maketitle

\begin{abstract}
Climate change's impact on human health poses unprecedented and diverse challenges. Unless proactive measures based on solid evidence are implemented, these threats will likely escalate and continue to endanger human well-being. The escalating advancements in information and communication technologies have facilitated the widespread availability and utilization of social media platforms. Individuals utilize platforms such as Twitter and Facebook to express their opinions, thoughts, and critiques on diverse subjects, encompassing the pressing issue of climate change. The proliferation of climate change-related content on social media necessitates comprehensive analysis to glean meaningful insights. This paper employs natural language processing (NLP) techniques to analyze climate change discourse and quantify the sentiment of climate change-related tweets. We use ClimateBERT, a pretrained model fine-tuned specifically for the climate change domain. The objective is to discern the sentiment individuals express and uncover patterns in public opinion concerning climate change. Analyzing tweet sentiments allows a deeper comprehension of public perceptions, concerns, and emotions about this critical global challenge. The findings from this experiment unearth valuable insights into public sentiment and the entities associated with climate change discourse. Policymakers, researchers, and organizations can leverage such analyses to understand public perceptions, identify influential actors, and devise informed strategies to address climate change challenges.
\end{abstract}


\section{Introduction}
According to the World Health Organization (WHO), the greatest health threat to people in the twenty-first century is climate change. Between 2030 and 2050, this risk is expected to cause an additional 250,000 fatalities annually and express itself in various ways. Many of these health concerns can be decreased or avoided with prompt and effective adaptation, but doing so necessitates in-depth research and policies that are multi-sectoral, multi-system, and collaborative at several scales\citep{fard2022climedbert}. One of the most important issues that needs significant attention is climate change. The majority of scientists agree that human activity is accelerating the Earth's climate change, which is having a disastrous effect on the world and its population. The consequences of climate change are more clear. In recent years, extreme weather occurrences like hurricanes, tornadoes, hail, lightning, fires, and floods have increased in frequency and intensity. As the world's ecosystems change quickly, access to the natural resources and agricultural methods that support humanity is in danger. \citep{ardabili2020deep}. The problem of climate change is complicated, and there is no quick fix. But to identify solutions, it's critical to comprehend the issue.
\par Large volumes of text data can be analyzed using natural language processing which may unearth interesting patterns\cite{anoop2023sentiment}. The natural language processing approaches can be applied to the climate change domain as well for finding the causes and leveraging patterns such as public sentiment and discourse towards this global issue. Recent years have witnessed many people using social media to share their views, concerns, and public opinions on any topic under the sky\cite{anoop2023we}\cite{jickson2023machine}. This has caused a huge amount of unstructured but dynamic data to be generated in such platforms, which are goldmines for social science researchers\cite{john2023health}\cite{anoop2023public}. Collecting, curating, and analyzing such data is crucial for finding public perceptions and viewpoints on socially relevant discussions\cite{varghese2022deep}\cite{lekshmi2022sentiment}. Similarly, understanding public opinions and sentiments on climate change is crucial for public policymakers, governments, and other administrators to devise better policies and intervention measures to address the challenges.
\par In this research, natural language processing is used to examine tweets that discuss climate change. We use ClimateBERT\cite{webersinke2021climatebert}, a pre-trained language model trained on a large set of climate change-related documents, and fine-tune the same for sentiment classification tasks. The findings may be used better to understand the public's understanding of climate change, and it can also be used to identify the key stakeholders in the climate change debate. The interesting insights from this project will provide a foundation for informed decision-making and policy formulation regarding climate change. Additionally, the findings will contribute to advancing NLP techniques and their application in climate change analysis. 
\subsection{Effects of Climate Change}
The consequences of climate change are extensive and have a significant impact on many facets of our planet. Several important aspects that shed information on the effects of climate change have been highlighted in research articles. Climate change is already manifesting globally, with a notable increase in extreme weather events. The frequency and intensity of hurricanes, floods, and droughts have amplified, causing widespread destruction and loss of life. Coastal areas are also seriously threatened by increasing sea levels, which might result in massive population displacement, increased erosion, and flooding. As glaciers continue to melt, water supplies diminish, affecting regions dependent on glacial meltwater for agricultural, industrial, and domestic purposes. Climate change-induced shifts in environmental conditions are causing profound changes in plant and animal life. Species are forced to adapt or face extinction as they grapple with altered ecosystems and changing habitats. This disruption to biodiversity has cascading effects on ecosystem functioning and services, with implications for food security, ecosystem stability, and human well-being.
\par Another consequence of climate change is the heightened risk of diseases. As temperatures rise, disease-carrying organisms, such as mosquitoes, expand their geographic range, exposing previously unaffected regions to vector-borne illnesses. This poses a significant public health challenge, necessitating the development of effective strategies for disease prevention, control, and surveillance. The impacts of climate change extend beyond the natural environment, affecting societies and economies globally. Disruptions to ecosystems and weather patterns have severe social and economic repercussions, with vulnerable communities being disproportionately affected. Climate-induced events, such as extreme heatwaves, prolonged droughts, and intense storms, lead to the displacement of populations, loss of livelihoods, and increased socioeconomic inequality. Consequently, countries face significant challenges in managing the economic and social ramifications of climate change, including the need for adaptation measures and the transition to sustainable practices. In summary, climate change is causing a range of effects that reverberate across multiple dimensions of our planet. The effects are widespread and provide serious difficulties for human society and ecosystems, from extreme weather events and increasing sea levels to melting glaciers and biodiversity loss. Addressing climate change requires concerted global efforts to mitigate greenhouse gas emissions, enhance resilience, and foster sustainable development practices to safeguard the future of our planet and its inhabitants.
\subsection{NLP for Climate Change Analysis}
Natural Language Processing (NLP) is an emerging discipline in computer science that concentrates on the creation of algorithms and models to facilitate computers in comprehending, analyzing, and producing human language. In the context of climate change analysis, NLP techniques have proven to be invaluable in extracting meaningful insights from vast amounts of textual data, offering a new perspective on this critical global issue. One application of NLP in climate change analysis is the ability to analyze public opinion on the topic. By leveraging sentiment analysis techniques, researchers can gauge the prevailing sentiments, attitudes, and beliefs surrounding climate change. This understanding of public opinion is crucial for policymakers, as it helps them tailor communication strategies, design effective interventions, and foster public engagement in addressing climate change challenges.
\par Furthermore, NLP techniques allow for identifying key stakeholders involved in the climate change debate.Through the extraction and examination of written information from various outlets, including news articles, social media platforms, and scientific journals, scholars can discern the key individuals, organizations, and institutions influencing the conversation surrounding climate change. This knowledge provides valuable insights into the various perspectives, interests, and motivations shaping climate change discussions, facilitating informed decision-making and targeted engagement with relevant stakeholders. NLP also enables the tracking of the progress of climate change negotiations. By analyzing texts from international agreements, policy documents, and meeting transcripts, researchers can monitor the evolution of climate change discussions, assess the effectiveness of existing frameworks, and identify areas of convergence or divergence among different stakeholders. This monitoring capability helps policymakers and negotiators evaluate the efficacy of climate change policies, identify potential barriers to progress, and inform future negotiations and policy development.
\par Additionally, NLP techniques can be applied to monitor the impact of climate change on different regions of the world. By analyzing textual data from scientific reports, environmental assessments, and socio-economic surveys, researchers can gain insights into the specific vulnerabilities, risks, and adaptation strategies associated with climate change in different geographic areas. This information is crucial for policymakers and local communities to prioritize resources, implement targeted interventions, and build resilience against the impacts of climate change. In summary, NLP techniques offer a powerful toolkit for analyzing climate change-related textual data, enabling researchers to gain valuable insights into public opinion, identify key stakeholders, track the progress of climate change negotiations, and monitor the impact of climate change on different regions. By harnessing the potential of NLP, policymakers and researchers can enhance their understanding of climate change dynamics and develop evidence-based strategies for mitigation, adaptation, and effective decision-making in the face of this global challenge.
The major contributions of this research may be summarized as follows:
\begin{itemize}
     \item Conducts a detailed study on different approaches reported in the natural language processing literature on sentiment analysis using social media data.
    \item Explores the potential of using ClimateBERT - a pre-trained model on climate data, for the sentiment analysis of tweets on climate change.
    \item Conducts extensive experiments and reports the experimental comparisons with different machine learning algorithms on sentiment analysis using ClimateBERT
\end{itemize}

\section{Related Studies}
This section provides an overview of recent and influential research papers in machine learning and natural language processing, specifically on climate change analysis. It also discusses relevant studies that explore sentiment text classification approaches that are pertinent to the proposed project. The reviewed studies have demonstrated the effectiveness of sentiment analysis and named entity recognition techniques in the context of climate change analysis. NLP models such as BERT and attention mechanisms have shown promising results in capturing contextual information and improving performance. These studies provide valuable insights and methodologies to guide our approach in implementing sentiment analysis and named entity recognition on climate change-related tweets and texts using the ClimateBERT pre-trained model.

A study was conducted to assess the effectiveness of ML algorithms in predicting long-term global warming. The research examined algorithms such as LR, SVR, lasso, and ElasticNet to connect average annual temperature and greenhouse gas factors. By analyzing a dataset spanning 100-150 years, the study found that carbon dioxide (CO2) had the most significant impact on temperature changes, followed by CH4, N2O, and SF6. Using this information, the researchers were able to forecast temperature trends and greenhouse gas levels for the next decade, providing valuable insights for mitigating the consequences of global warming. The research analyzes public sentiments regarding climate change by studying Twitter data. The study aims to tackle the problems of polarization and misinformation that often arise during climate change discussions on social media platforms. To achieve this, the researchers introduce a multi-task model named MEMOCLiC, which combines stance detection with additional tasks like emotion recognition and offensive language identification. By employing various embedding techniques and attention mechanisms, the proposed framework effectively captures specific characteristics and interactions related to different modalities. Experimental findings highlight the superior performance of the MEMOCLiC model in enhancing stance detection accuracy compared to baseline methods.
\par This research paper examines the issue of polarization and belief systems prevalent in climate change discussions on Twitter. The paper proposes a framework that aims to identify statements denying climate change and classify tweets into two categories: denier or believer stances. \citep{sham2022climate}The framework focuses on two interconnected tasks: stance detection and sentiment analysis. Combining these tasks, the multi-task model utilizes feature-specific and shared-specific attention frameworks to acquire comprehensive features. Experimental results demonstrate that the proposed framework enhances stance detection accuracy by leveraging sentiment analysis, outperforming uni-modal and single-task approaches. This research paper utilizes the BERT model and convolutional neural network (CNN).\citep{lydiri2022performant} The study analyzes public opinions on climate change by examining Twitter data. The results indicate that the proposed model surpasses conventional machine learning methods, accurately identifying climate change believers and deniers. The authors suggest this model has significant potential for monitoring and governance purposes, particularly in smart city contexts. Additionally, future work involves investigating alternative deep learning algorithms and expanding the analysis to encompass other social media platforms.
\par This research paper \citep{ceylan2022application}investigates the application of AI and NLP models to analyze extensive unstructured data concerning climate change. The study primarily aims to develop an information management system capable of extracting pertinent information from diverse data sources, particularly technical design documentation. By utilizing pre-trained AI-based NLP models trained on textual data and integrating non-textual graphical data, the researchers showcase the system's effectiveness in swiftly and efficiently retrieving precise information. The ultimate objective is to promote knowledge democratization and ensure the accessibility of information to a broad user base. This research paper examines people's emotions and opinions concerning the conflict between Russia and Ukraine by employing ML and DL techniques. \citep{sirisha2022aspect}The study introduces a novel hybrid model combining sequence and transformer models, namely ROBERTa, ABSA, and LSTM. To conduct the analysis, a large dataset of geographically tagged tweets related to the Ukraine-Russia war is collected from Twitter, and sentiment analysis is performed using the proposed model. The findings indicate that the hybrid model achieves a remarkable accuracy of 94.7, surpassing existing approaches in sentiment analysis. The study underscores the significance of social media platforms such as Twitter in gaining insights into public sentiment and opinions regarding global events.
\par This research paper aims to overcome the limitations of general language models in effectively representing climate-related texts. The authors introduce CLIMATEBERT, a transformer-based language model that undergoes pretraining on a vast corpus of climate-related paragraphs extracted from diverse sources such as news articles, research papers, and corporate disclosures.\citep{webersinke2021climatebert} Comparative evaluations reveal that CLIMATEBERT surpasses commonly used language models, exhibiting a substantial 48 enhancement in a masked language model objective. The improved performance of CLIMATEBERT contributes to lower error rates in various climate-related downstream tasks. To encourage further research at the intersection of climate change and natural language processing, the authors provide public access to the training code and weights of CLIMATEBERT. This research paper uses ML algorithms to analyze and predict climate change. The authors emphasize the significance of comprehending and adapting to the impacts of climate change on both human society and the environment. The study discusses the application of ML methods in analyzing historical temperature data and carbon dioxide concentrations dating back to the 18th century. It emphasizes the potential advantages of employing machine learning and artificial intelligence in interpreting and harnessing climate data for simulations and predictions. Multiple machine learning algorithms, such as DT, RF, and ANN, are examined for climate change risk assessment and prediction. The authors conclude that integrating machine learning techniques can enhance climate modeling, enabling informed decision-making concerning climate change mitigation and adaptation strategies.
\par This research paper introduces the ClimaText dataset\citep{varini2020climatext}, specifically developed to detect sentence-level climate change topics within textual sources. The authors emphasize the significance of automating the extraction of climate change information from media and other text-based materials to facilitate various applications, including content filtering, sentiment analysis, and fact-checking. Through a comparative analysis of different approaches for identifying climate change topics, they find that context-based algorithms like BERT outperform simple keyword-based models. However, the authors also identify areas that require improvement, particularly in capturing the discussion surrounding the indirect effects of climate change. The authors anticipate this dataset will be a valuable resource for further research in natural language understanding and climate change communication. \citep{upadhyaya2022multitask}It underscores the importance of comprehending public perception and acceptance of climate change policies. The study examines diverse data sources, such as social media, scientific papers, and news articles, to perform sentiment analysis. ML techniques, specifically SVM, are evaluated for extracting valuable insights from these data sources. The paper concludes that supervised machine learning techniques exhibit effectiveness in sentiment analysis, highlighting that ensemble and hybrid approaches yield superior outcomes compared to individual classifiers.
\section{Materials and Methods}
\subsection{Label Studio}
Label Studio is an open-source data annotation tool (available at \url{https://labelstud.io/}) that provides a user-friendly interface for creating labeled datasets by annotating data for machine learning and artificial intelligence tasks. The tool supports various annotation types, including text classification, NER, object detection, image segmentation, and more. Label Studio allows users to import data from various sources, such as CSV files, JSON, or databases, and annotate them using a customizable interface. It provides a collaborative environment where multiple annotators can collaborate on a project, with features like task assignment, annotation review, and inter-annotator agreement measurement. One of the key features of Label Studio is its extensibility. It provides a flexible architecture that allows users to customize the annotation interfaces and incorporate custom labeling functions using JavaScript and Python. This enables the tool to adapt to different annotation requirements and integrate with existing machine-learning workflows. Label Studio also supports active learning, where the tool can suggest samples to be annotated based on a model's uncertainty, helping to optimize the annotation process and improve model performance.
\subsection{snscrape}
\textit{\textbf{snscrape}} is a Python library and command-line tool (available at \url{https://github.com/JustAnotherArchivist/snscrape}) for scraping social media content. It lets you retrieve public data from various social media platforms, including Twitter, Instagram, YouTube, Reddit, etc. With snscrape, you can fetch posts, comments, likes, followers, and other relevant information from social media platforms. It provides a flexible and customizable way to search for specific keywords, hashtags, usernames, or URLs and extract the desired content. The library supports scraping recent and historical data from social media platforms, enabling you to gather insights, perform analysis, monitor trends, and conduct research based on social media content. snscrape offers a command-line interface that allows you to search for and scrape social media data interactively. You can specify various parameters, such as the number of results, date range, and output format, to customize your scraping process. In addition to the command-line interface, snscrape provides a Python API that allows you to integrate social media scraping into your own Python scripts and applications. The API offers more advanced functionalities, giving you fine-grained control over the scraping process and allowing you to process the scraped data programmatically. One of the key advantages of snscrape is its ability to work with multiple social media platforms, providing a unified interface for scraping different types of content. It handles the intricacies of each platform's APIs and HTML structures, making it easier for developers to extract data without needing to learn the specific details of each platform. It's important to note that snscrape respects the terms of service and usage restrictions of each social media platform. It is primarily intended for scraping publicly available content and should be used responsibly and in compliance with the platform's policies.
\subsection{Newspaper 3k}
Newspaper3k is a Python library and web scraping tool (available at \url{https://newspaper.readthedocs.io/}) that allows you to extract and parse information from online news articles. It provides a simple interface to automate the fetching and processing of news articles from various online sources. With Newspaper3k, you can retrieve article metadata such as the title, author, publish date, and article text from news websites. It also supports extracting additional information like keywords, summaries, and article images. The library uses advanced NLP techniques to extract relevant information from the HTML structure of the news articles. Newspaper3k is designed to handle various complexities of news websites, including different article formats, pagination, and content extraction. It has built-in functionality to handle newspaper-specific features like multi-page articles, article pagination, and RSS feeds. One of the advantages of Newspaper3k is its ease of use. It abstracts away the complexities of web scraping and provides a clean and intuitive API. It also handles various encoding and parsing issues that often arise when dealing with news articles from different sources. Newspaper3k is widely used for various applications, including content analysis, sentiment analysis, and data mining. It offers a convenient way to gather news data for research, data analysis, and machine learning projects.
\subsection{ClimateBERT}
ClimateBERT is a specialized variant of the BERT model specifically trained and tailored for addressing climate change-related language tasks. Building upon the foundation of BERT, ClimateBERT is pre-trained on a large corpus of climate change-related documents and text sources, enabling it to capture the nuances and domain-specific knowledge relevant to climate science. This fine-tuning process equips ClimateBERT with a deep understanding of climate-related concepts, terminology, and contextual dependencies. \citep{iqbal2023omicron} By leveraging ClimateBERT, researchers and practitioners in climate change analysis can effectively tackle various NLP tasks, such as sentiment analysis on climate-related tweets or named entity recognition on climate change articles. Integrating domain-specific knowledge into the pre-training process makes ClimateBERT a powerful tool for extracting insights, identifying patterns, and extracting valuable information from climate-related text data. Its application in climate change analysis can aid in improving decision-making, facilitating research, and enhancing our understanding of the complex challenges climate change poses.
\section{Proposed Approach}
This section deals with the proposed methodology for the sentiment analysis of climate-related tweets from Twitter using ClimateBERT embeddings and
Random Forest Classifier. The overall workflow of the proposed approach is given in Figure 1.
\begin{figure}[htb]
\centering
\includegraphics[width=12cm]{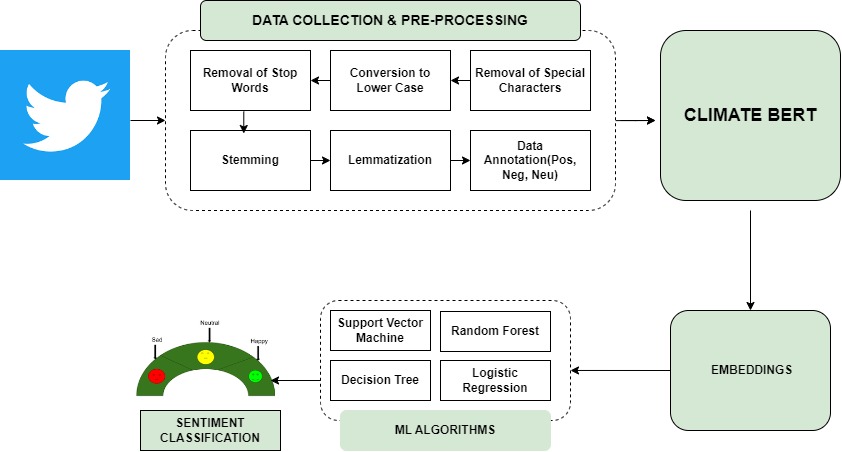}
\caption{Overall workflow of the sentiment analysis on climate Change tweets}
\end{figure}
\subsection{Dataset}
The methodology begins with data collection from Twitter using the snscrape library. The collected data is loaded into a pandas DataFrame for further processing. The dataset consists of climate change-related tweets, valuable for sentiment analysis. The tweets gathered between 1 January 2022 and 2 February 2023. However, it's important to note that the collected data may contain class imbalance, where certain sentiment categories are overrepresented while others are underrepresented. This could potentially bias the model's predictions. Initially, the dataset consisted of 4410 data points. After data augmentation, the final dataset consists of 5506 data points, with three labels, Positive, Negative and Neutral. The data set is available at \url{https://github.com/appliednlp-duk/nlp-climate-change}. Table 1 shows the example of labeling for each tweet, whether it is Positive, Negative, or Neutral.
\begin{table}[!ht]
\caption{A snapshot of the dataset used for the experiment}
    \centering
    \begin{tabular}{|l|l|}
    \hline
        Content & Labels \\ \hline
        \parbox{8cm} {Researchers use deep learning to simulate chlorophylla \& phycocyanin with an internet of things system to detect \& \\quantify \#cyanobacteria, to improve \#eutrophication management schemes for freshwater reservoirs.\\
\#algae \#microbiology \#environment \#iot 
eeer.org/journal/view.pâ€¦ } & Positive  \\ \hline
        \parbox{8cm} {Why is our @Conservatives government so evil?

\#RishiSunak
\#climateChange \#Conservatives
\#FuckingThieves https://t.co/cCGyylmYlf } & Negative \\ \hline
        \parbox{8cm} {Sierra snowpack 205\% of its historical average | Climate Change ... - San Francisco Examiner dlvr.it/ShpGVN \#ClimateChange } & Neutral  \\
\hline
    \end{tabular}
\end{table}

\subsection{Data Preprocessing}
To prepare the data for sentiment analysis, several preprocessing steps are applied. Special characters and digits are removed, and the text is converted to lowercase to ensure consistency. Tokenization is performed, which involves splitting the words into individual units. Stopwords (common words with little contextual meaning) are removed, and stemming or lemmatization techniques may be applied to normalize the words. This preprocessing step ensures the text data is cleaned and ready for further analysis.
\subsection{Experimental setup}
This section describes the experiment for implementing the proposed approach detailed in section 3. All the experiments were executed on NVIDIA A100 with 80 GB GPU memory and 1,935 GB/Second bandwidth. The ChatGPT Sentiment tweets were pre-processed to make them ready for experimentation. All scripts were written in Python 3.9, and the Machine Learning Models were used from the Scikit-Learn library available at \url{https://scikit learn.org/stable/}.
\section{Results and Discussions}
This section presents the results obtained from the experiment using the proposed approach outlined in Section 4. The results, along with a detailed discussion, are given in this section.
\begin{table*}[ht!]
\centering
\caption{Precision, Recall, Accuracy, and F-Measure values for the TF-IDF feature encoding}
\begin{tabularx}{.9\textwidth}{lllllllll}
\toprule
Model & \multicolumn{4}{c}{TF-IDF}        \\
      & Accuracy     & Precision     & Recall     & F-measure          \\
\midrule
RF   & 87.11 & 86.98 & 87.11 & 87.01  \\
SVM   & 84.39  & 84.34 & 84.39  & 84.14 &  \\
DT    & 73.23  & 72.63 & 73.23  & 72.71   \\
LR   & 90.10 & 90.13 & 90.10 & 90.04  \\
\bottomrule
\end{tabularx}
\end{table*}

\begin{table*}[ht!]
\centering
\caption{Precision, Recall, Accuracy, and F-Measure values for word2Vec}
\begin{tabularx}{.9\textwidth}{lllllllll}
\toprule
Model & \multicolumn{4}{c}{word2Vec}        \\
      & Accuracy     & Precision     & Recall     & F-measure          \\
\midrule
RF  & 72.59 & 72.91 & 72.59 & 72.69  \\
SVM   & 40.29  & 40.59 & 40.29  & 36.92 &  \\
DT   & 62.34  & 61.24 & 62.34  & 61.59   \\
LR   & 39.56 & 40.25 & 39.56 & 36.53  \\
\bottomrule
\end{tabularx}
\end{table*}

\begin{table*}[ht!]
\centering
\caption{Precision, Recall, Accuracy, and F-Measure values for CountVectorizer}
\begin{tabularx}{.9\textwidth}{lllllllll}
\toprule
Model & \multicolumn{4}{c}{CountVectorizer}        \\
      & Accuracy     & Precision     & Recall     & F-measure          \\
\midrule
RF  & 89.56 & 89.53 & 89.56 & 89.50  \\
SVM    & 88.11  & 88.02 & 88.11  & 87.98 &  \\
DT   & 79.67  & 79.50 & 79.67  & 79.58   \\
LR   & 79.21 & 79.09 & 79.21 & 78.76  \\
\bottomrule
\end{tabularx}
\end{table*}

\begin{table*}[ht!]
\centering
\caption{Precision, Recall, Accuracy, and F-Measure values for TF-IDF + CountVectorizer}
\begin{tabularx}{.9\textwidth}{lllllllll}
\toprule
Model & \multicolumn{4}{c}{TF-IDF + CountVectorizer}        \\
      & Accuracy     & Precision     & Recall     & F-measure          \\
\midrule
RF   & 86.93 & 86.96 & 86.93 & 86.87  \\
SVM  & 81.94  & 81.92 & 81.94  & 81.75 &  \\
DT   & 74.77  & 74.30 & 74.77  & 74.46   \\
LR  & 81.30 & 81.11 & 81.30 & 80.99  \\
\bottomrule
\end{tabularx}
\end{table*}

\begin{table*}[ht!]
\centering
\caption{Precision, Recall, Accuracy, and F-Measure values for TF-IDF + word2Vec}
\begin{tabularx}{.9\textwidth}{lllllllll}
\toprule
Model & \multicolumn{4}{c}{TF-IDF + word2Vec}        \\
      & Accuracy     & Precision     & Recall     & F-measure          \\
\midrule
RF   & 79.21 & 79.45 & 79.21 & 79.29  \\
SVM    & 85.39  & 85.34 & 85.39  & 85.22 &  \\
DT   & 62.43  & 61.28 & 62.43  & 61.60   \\
LR    & 74.77 & 74.33 & 74.77 & 74.28  \\
\bottomrule
\end{tabularx}
\end{table*}

\begin{table*}[ht!]
\centering
\caption{Precision, Recall, Accuracy, and F-Measure values for CountVectorizer + word2Vec}
\begin{tabularx}{.9\textwidth}{lllllllll}
\toprule
Model & \multicolumn{4}{c}{CountVectorizer + word2Vec}        \\
      & Accuracy     & Precision     & Recall     & F-measure          \\
\midrule
RF   & 79.03 & 79.45 & 79.03 & 79.18  \\
SVM    & 81.30  & 81.10 & 81.30  & 81.09 &  \\
DT  & 62.06  & 61.01 & 62.06  & 61.37   \\
LR   & 81.21 & 81.04 & 81.21 & 80.92  \\
\bottomrule
\end{tabularx}
\end{table*}

\begin{table*}[htb]
\centering
\caption{Performance evaluation of SVM, LR, RF, and DT algorithms using BERT}
\begin{tabularx}{.9\textwidth}{lllllllll}
\toprule
Model & \multicolumn{4}{c}{BERT}        \\
      & Accuracy     & Precision     & Recall     & F-measure          \\
\midrule
RF   & 76.78 & 77.46 & 76.78 & 76.93  \\
SVM    & 64.35  & 63.65 & 64.35  & 63.70 &  \\
DT    & 68.89  & 67.13 & 68.89  & 67.89   \\
LR    & 63.81 & 63.48 & 63.81 & 63.60  \\
\bottomrule
\end{tabularx}
\end{table*}

\begin{table*}[htb]
\centering
\caption{Performance evaluation of SVM, LR, RF, Naive Bayes, and DT algorithms using ClimateBERT}
\begin{tabularx}{.9\textwidth}{lllllllll}
\toprule
Model & \multicolumn{4}{c}{ClimateBERT}        \\
      & Accuracy     & Precision     & Recall     & F-measure          \\
\midrule
\textbf{RF}   & \textbf{85.22} & 85.73 & 85.22 & 83.33 \\
SVM   & 75.66  & 76.20 & 75.66  & 75.07 &  \\
DT    & 80.62  & 79.88 & 78.62  & 77.47   \\
LR    & 73.84 & 72.92 & 73.84 & 75.69  \\
\bottomrule
\end{tabularx}
\end{table*}

\begin{figure}
     \centering
     \begin{subfigure}[b]{0.4\textwidth}
         \centering
         \includegraphics[width=\textwidth]{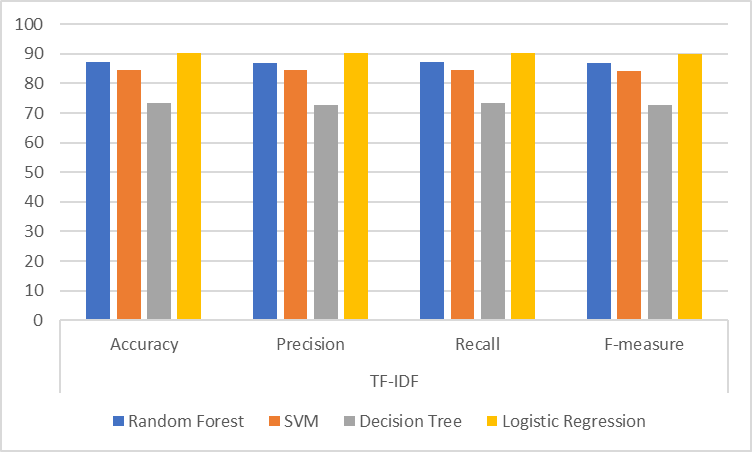}
         \caption{Performance evaluation of SVM, LR, RF, and DT algorithms with TF-IDF}
     \end{subfigure}
     \hfill
     \begin{subfigure}[b]{0.4\textwidth}
         \centering
         \includegraphics[width=\textwidth]{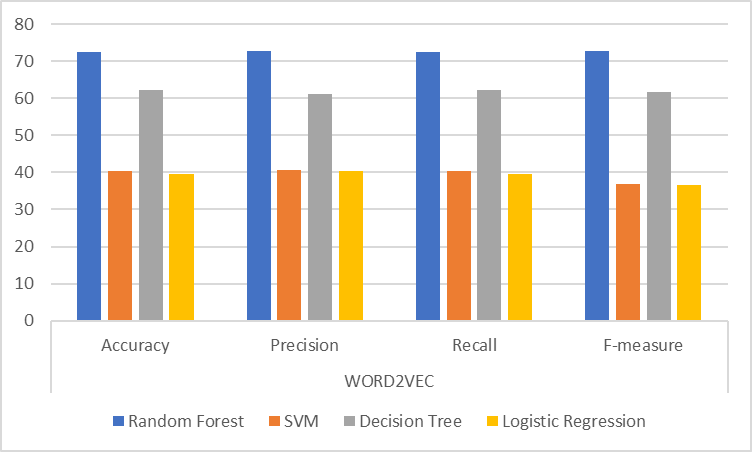}
         \caption{Performance evaluation of SVM, LR, RF, and DT algorithms with TF-IDF}
     \end{subfigure}
\end{figure}

\begin{figure}
     \centering
     \begin{subfigure}[b]{0.4\textwidth}
         \centering
         \includegraphics[width=\textwidth]{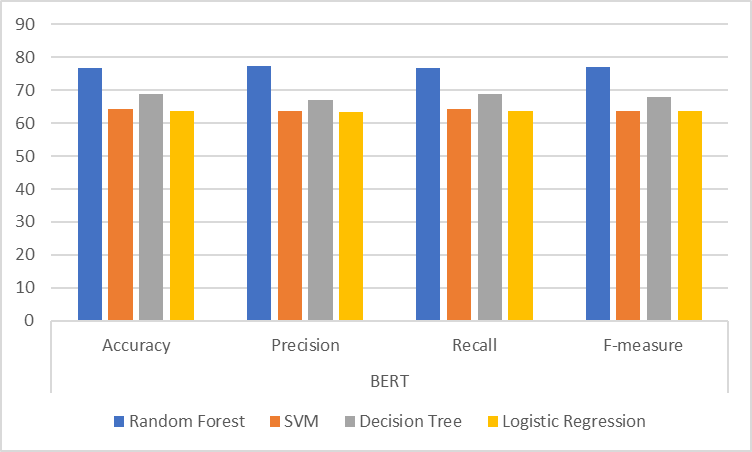}
         \caption{Performance evaluation of SVM, LR, RF, and DT algorithms with BERT}
     \end{subfigure}
     \hfill
     \begin{subfigure}[b]{0.4\textwidth}
         \centering
         \includegraphics[width=\textwidth]{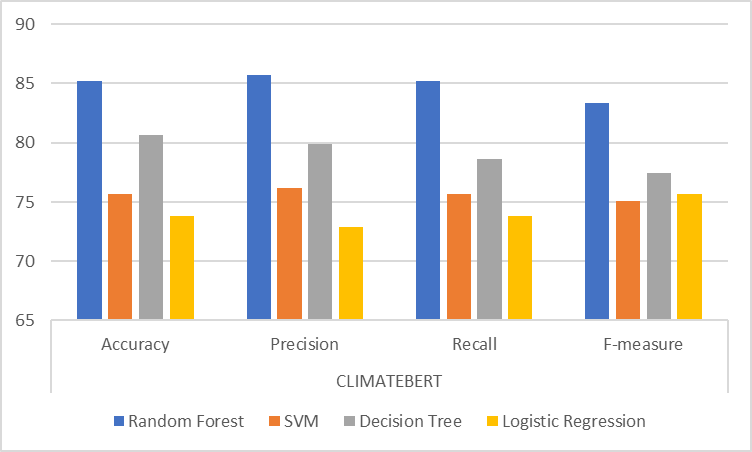}
         \caption{Performance evaluation of SVM, LR, RF, and DT algorithms with ClimateBERT}
     \end{subfigure}
\end{figure}

\begin{figure}
     \centering
     \begin{subfigure}[b]{0.4\textwidth}
         \centering
         \includegraphics[width=\textwidth]{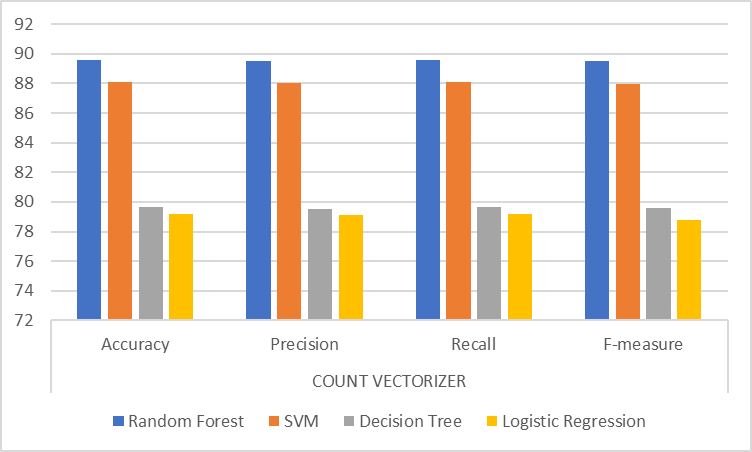}
         \caption{Performance evaluation of SVM, LR, RF, and DT algorithms with CountVectorizer}
     \end{subfigure}
     \hfill
     \begin{subfigure}[b]{0.4\textwidth}
         \centering
         \includegraphics[width=\textwidth]{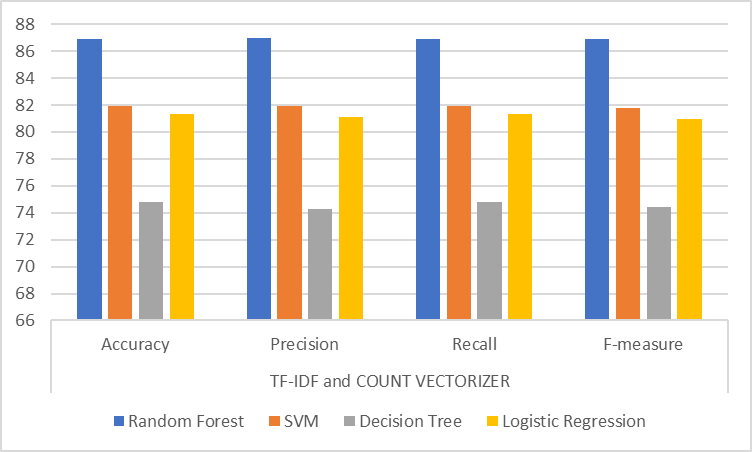}
         \caption{Performance evaluation of SVM, LR, RF, and DT algorithms with Combination of TF-IDF and CountVectorizer}
     \end{subfigure}
\end{figure}

\begin{figure}
     \centering
     \begin{subfigure}[b]{0.4\textwidth}
         \centering
         \includegraphics[width=\textwidth]{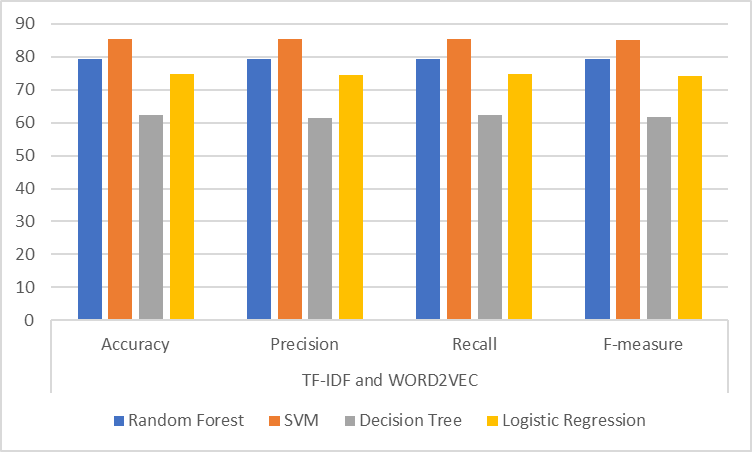}
         \caption{Performance evaluation of SVM, LR, RF, and DT algorithms with Combination of TF-IDF and Word2Vec}
     \end{subfigure}
     \hfill
     \begin{subfigure}[b]{0.4\textwidth}
         \centering
         \includegraphics[width=\textwidth]{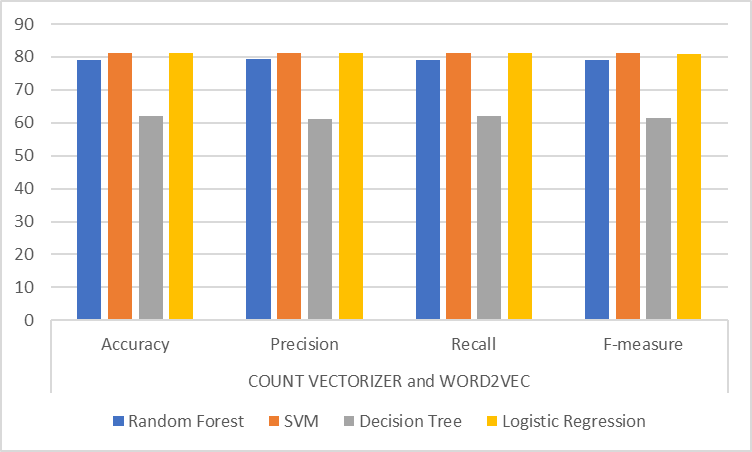}
         \caption{Performance evaluation of SVM, LR, RF, and DT algorithms with Combination of CountVectorizer and Word2Vec}
     \end{subfigure}
\end{figure}

Table 1 to 8 shows the accuracy, precision, recall, and f-measure values for RF, SVM, DT, and LR algorithms. For BERT embeddings, RF has 76.78\%, 77.46\%, 76.78\%, and 76.93\%, SVM has 64.35\%, 63.65\%, 64.35\%, and 63.70\%, DT has 68.89\%, 67.13\%, 68.89\%, and 67.59\%, and LR has 63.81\%, 63.48\%, 63.81\%, and 63.60\% for the A, P, R, and F values. Table 4 shows the Accuracy, Precision, Recall, and F-measure values for RF, SVM, DT, and LR algorithms. For ClimateBERT embeddings, RF has 85.22\%, 85.73\%, 85.22\%, and 83.33\%, SVM has 75.66\%, 76.20\%, 75.66\%, and 75.07\%, DT has 80.62\%, 79.88\%, 78.62\%, and 77.47\%, and LR has 73.84\%, 72.92\%, 73.84\%, and 75.69\% for the A, P, R, and F values. After training, the model's performance is evaluated on the test set to assess its ability to predict sentiment. The model is switched to evaluation mode, and predictions are made on the test set. Accuracy, precision, recall, and F1-score are calculated to measure the model's performance. The results obtained from the evaluation metrics are reported. Accuracy provides an overall measure of correctness, precision measures the proportion of correctly predicted positive sentiments, recall captures the ability to identify all positive sentiments, and the F1-score provides a balanced measure between precision and recall. These metrics provide insights into how well the model predicts sentiment on climate change-related tweets. By following this experimental setup, the methodology ensures that the collected data is cleaned, balanced, and used effectively to train a sentiment analysis model. The results and discussions provide valuable insights into the model's performance and its ability to analyze sentiment in climate change discussions on Twitter.
\section{Conclusions}
Climate change, a pressing global concern, necessitates thorough analysis and understanding across diverse domains to mitigate its impacts effectively. In recent years, the fusion of NLP techniques and machine learning algorithms has emerged as a promising approach for comprehending the complexities and nuances of climate change through the lens of textual data. This paper utilized the advancements in domain-specific large language models to harness the potential of NLP in addressing the challenges posed by climate change through sentiment analysis. By leveraging advanced NLP methodologies, we could identify climate change discourse that may enable uncovering valuable insights and facilitate informed decision-making.
\bibliographystyle{plainnat}
\bibliography{references}  
\end{document}